\documentclass[letterpaper, 10 pt, conference]{ieeeconf}  

\IEEEoverridecommandlockouts                              

\usepackage{graphics} 
\usepackage{epsfig} 

\usepackage {algorithm}
\usepackage {algorithmic}

\title{\LARGE \bf
Avalon: Building an Operating System for Robotcenter
}

\author{
Yuan Xu, Zhiyuan Yan, Sa Wang, Cheng Yang, Qingsai Xiao and Yungang Bao
\thanks{The authors are with the State Key Laboratory of Computer Architecture, Institute of Computing Technology, Chinese Academy of Science.
The contact author is Y. Xu, e-mail: xuyuan@ict.ac.cn}
}

\begin{document}

\maketitle
\thispagestyle{empty}
\pagestyle{empty}

\begin{abstract}

This paper envisions a scenario that hundreds of heterogeneous robots form a robotcenter which can be
shared by multiple users and used like a single powerful robot to perform complex tasks.
However, current multi-robot systems are either unable to manage heterogeneous robots or
unable to support multiple concurrent users. Inspired by the design of modern datacenter OSes, 
we propose Avalon, a robot operating system with two-level scheduling scheme which is widely
adopted in datacenters for Internet services and cloud computing. Specifically, Avalon integrates 
three important features together: (1) Instead of allocating a whole
robot, Avalon classifies fine-grained robot resources into three categories to distinguish
which fine-grained resources can be shared by multi-robot frameworks simultaneously. (2) Avalon adopts
a location based resource allocation policy to substantially reduce scheduling overhead. 
(3) Avalon enables robots to offload computation intensive tasks to the clouds. 
We have implemented and evaluated Avalon on robots on both simulated environments and real world. 
\end{abstract}

\section{INTRODUCTION}

In this paper we ask the following question: can warehouse-scale heterogeneous robots
in a specific environment be managed to form a shared resource platform offered to multiple users?
We call such facilities as \textbf{robotcenters}, which are similar to
datacenters that provide resources on demand for cloud users.
For instance, in a modern office building, hundreds of robots purchased from different companies 
are busying with their own tasks. The robots are shared by all users in the building to 
cooperatively complete users' requests. Besides, they can simultaneously perform operations
for environment detection and monitoring, such as face recognition, 
position detection and behaviors analysis.

To deploy such a robotcenter, an operating system (OS) is required to manage both robot resources 
and user requests. In particular, we argue that future robotcenter OS needs to have the following features:
1) it supports heterogeneous robots with different shapes and functionalities; 
2) it supports flexible scalability and fast deployment; 
3) it supports fine-grained resource sharing so that one robot can be shared by multiple tasks simultaneously;
4) it allows multiple multi-robot frameworks to be deployed on a group of shared robots; 
5) it allows robots to offload computation to cloud-side datacenters. 

Over the recent years, there have been efforts on multi-robot frameworks that can coordinate a team of robots
to complete a specific task such as drawing on the sky. Unfortunately, they are unable to fulfill the
above features so as not to be suitable for robotcenters yet.
Specifically, current researches on multi-robot frameworks like Kiva \cite{Kiva} 
and Swarmfarm \cite{Swarmfarm} are designed for a specific task in a specific environment rather than
being shared by multiple users. In addition, these frameworks mainly aim to coordinate homogeneous 
robots \cite{Heterogeneity}, while robotcenters can consist of highly heterogeneous robots including 
Unmanned Ground Vehicle (UGV), Unmanned Aerial Vehicle (UAV) and even Unmanned Underwater Vehicle (UUV).

\begin{figure}[tb]
\centering
\includegraphics[width=8cm]{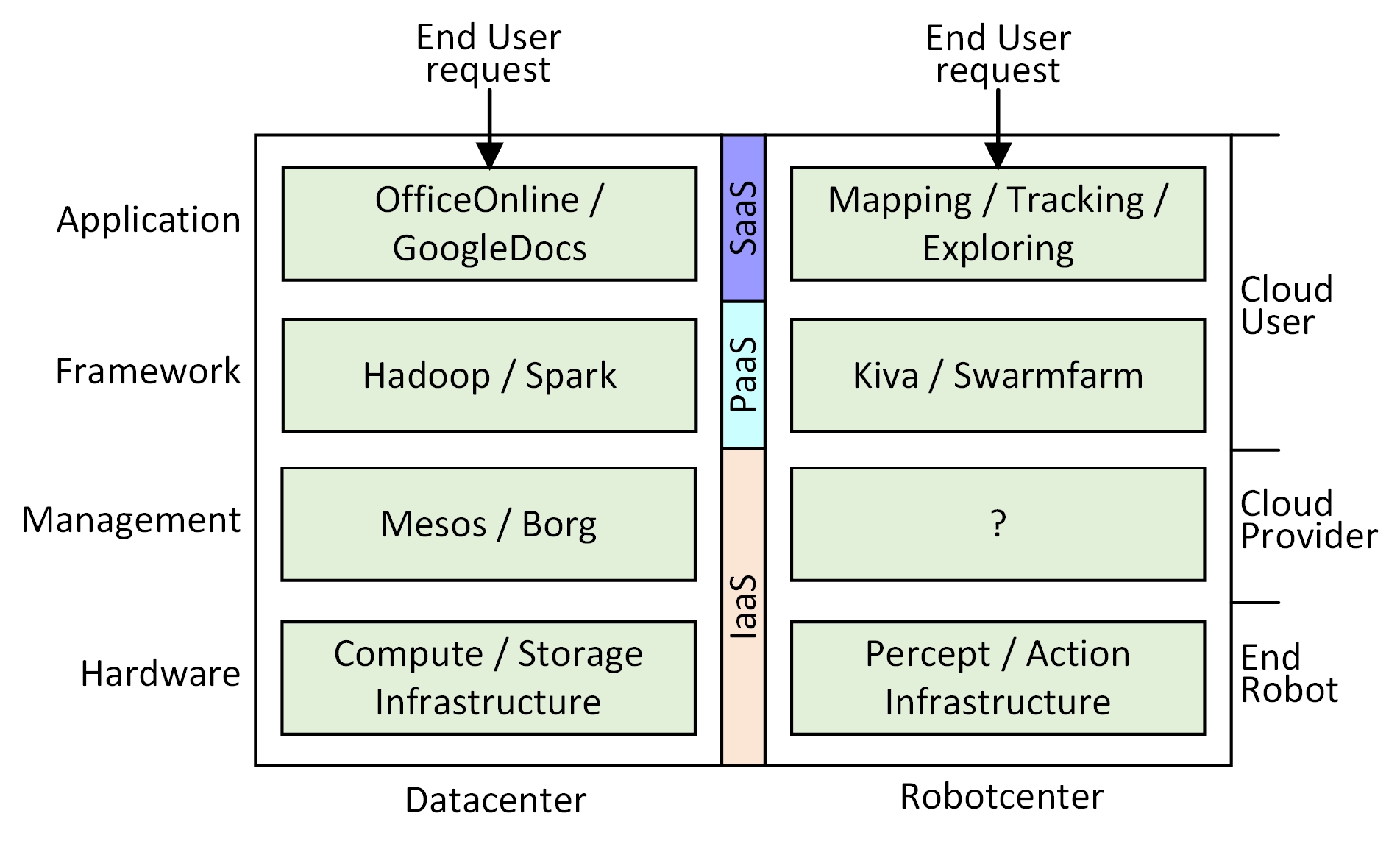}
\caption{Comparison between Robotcenter and Datacenter: Each robot connected to cloud will be
orchestrated as a resource pool with clusters, which allocated to a diverse range of multi-robot frameworks
for specific robotics application. }
\label{fig:robotcenter}
\end{figure}

Recently, the RoboEarth project \cite{RoboEarth} has made a substantial
contribution towards managing a group of robots. RoboEarth proposes a centralized task
controller for multiple robots and enables offloading computation to cloud 
by introducing Rapyuta \cite{Rapyuta}. However, it encounters a main limitation:
RoboEarth treats a whole robot as the minimal resource allocation unit. 
Thus, robots are managed in a batch mode and multiple jobs are sequentially 
executed on a robot. This mode is just like cluster-based supercomputer’s
job scheduling systems that treat a whole server 
as the minimal resource allocation unit. Nevertheless, it is well known that
batch-mode management is unsuitable for interactive jobs and usually results
in poor utilization. The advent of fine-grained resource management systems 
such as Borg \cite{Borg}, Kubernetes \cite{Kubernetes} and Mesos \cite{Mesos} significantly change 
the supercomputer-style’s cluster operating systems
and has established a new ecosystem of system software stack for 
modern datacenters. Today, it is common for a typical datacenter to serve mixed workloads including 
latency-sensitive interactive applications and batch applications.

As shown in Figure \ref{fig:robotcenter}, we observe that robotcenter and datacenter
exhibit similar system stack. Much like the multi-robot frameworks (e.g. Kiva, Swarmfarm), 
there are also many computing frameworks (e.g. Hadoop \cite{Hadoop}, Spark \cite{Spark} 
and TensorFlow \cite{Tensorflow}) in datacenters with different execution models such as 
MapReduce \cite{MapReduce} and Bulk-Synchronous-Parallel (BSP)) \cite{BSP}. 
To allow these frameworks to simultaneously share datacenter infrastructure, 
prevalent datacenter OSes such as Borg, Kubernetes and Mesos leverage
a two-level scheduling mechanism to fulfill fine-grained sharing for multiple 
computing frameworks. Datacenter Oses decouple resource allocation and 
task scheduling into two levels: the lower level is responsible for 
only fine-grained resource allocation and the higher level 
is responsible for task scheduling that is delegated to computing 
frameworks. We find that this mechanism can also be adopted for robotcenter OS.

Based on the observation, we propose a robotcenter OS design -- Avalon. To our best, it is 
the first system that applies two-level scheduling for managing multiple robots.
Specifically, Avalon allocates fine-grained robot resources to different multi-robot frameworks 
and delegates control over scheduling to the task allocator of each multi-robot framework.
Achieving this is challenging for two key reasons:
\begin{itemize}
  \item Unlike computing and storage resources in stationary servers,
  a robot's resources have different characteristics and not all of them can be shared
  by multiple frameworks. Therefore, a careful resource classification is required for
  distinguishing \emph{which} resource can be shared so as to enable
  multiplexing robots among multi-robot frameworks to improve robotcenter utilization.
  \item Prior multi-robot systems usually pre-allocate robots for a specific known task.
  However, like datacenters, robots in a robotcenters can be allocated on-demand and dynamically, 
  which raises a significant challenge for robot resource allocation.
\end{itemize}

To address these challenges, We make the following contributions:
\begin{itemize}
  \item \emph{Fine-grained resource sharing}. We classify robot resources into three categories: \emph{computation resources (CR)},
  \emph{sensory resources (SR)} and \emph{action resources (AR)}. This classification, 
  allows a robotcenter OS to know \emph{which} resource can be shared and \emph{how} frameworks use them.
  Based on the knowledge, we propose an efficient resource allocation policy to share CR and SR across multi-robot frameworks.
  \item \emph{Location-based resource offers}. To address the allocation on-demand challenge, 
  datacenter OSes usually use the locality principle to improve resource allocation. In contrast, 
  Avalon adopts a location-based resource management method. Specifically, we design a location
  based resource allocator that uses a cost function based on the distance between an operation's position
  and each robot's position. We also design a range \emph{filter} to reduce the number of resource offer list.
  By doing so, Avalon allows multi-robot frameworks to schedule without considering spatial cost,
  thereby resulting good system scalability.
  \item \emph{Open source prototype.} To demonstrate the feasibility of our design, we implemented Avalon 
  based on a datacenter OS Mesos \cite{Mesos} which supports Linux 4.4.0 and ROS indigo. It is worth noting
  that Avalon supports robots to offload their computation to the cloud, which is a promising direction for 
  the future Cloud-Edge computing mode \cite{EdgeComputing}.  
  Furthermore, due to the compatibility with ROS, Avalon can leverage more than 3000 open source packages to help 
  developers build multi-robot frameworks. We will also open source Avalon\footnote{A public release of Avalon is 
  available from https://github.com/xuyuan-ict/robotcenter}. 
\end{itemize}

  We evaluate Avalon on a Gazebo-simulated multi-robot system and on a turtlebot robot in real world.
  Experimental results show that Avalon is able to enable one robot to be shared by multiple tasks
  and improves CPU utilization average by 3.68 times (up to 7.51 times).


\section{BACKGROUND AND MOTIVATION}
Traditional multi-robot frameworks usually treat a whole robot as resource allocation unit. 
In fact, different applications pose different resource demands. Thus in this section, 
we will conduct experiments to demonstrate that one robot has the potential 
to be simultaneously shared by multiple applications,
which implies the feasibility of our proposed robotcenter OSes.

Assume that there are three multi-robot frameworks, each of which consists of a data-driven analysis task, 
a machine learning task and a SLAM task respectively. More description on these tasks are as below: 
\begin{itemize}
  \item \textbf{Data-driven analysis task}: This task regards robots as mobile sensors。 
  It collects perception data from each robot by accessing sensory resources \cite{RS1, RS2}. 
  In our experiments, we choose a monitoring workload, 
  which receives image data of a turtlebot robot and sends the data to cloud server continuously.
  \item \textbf{Machine learning task}: This task requires strong real-time constraints and 
  a lot of computing resources. It improves robots' intelligence by learning knowledge from sensory data about environments.
  In our experiment, we choose an image recognition workload that consists of a 34-layer convolutional 
  neural network (CNN) called resnet34 \cite{Resnet}, which offloads on-line training part to the backend cloud servers 
  and only performs inference on the robot.
  \item \textbf{SLAM task}: This task drives robots to execute actions in the physical world. 
  It takes in data from sensor streams and outputs executing commands to each actuator.
  We choose gmapping, a laser based SLAM workload, which uses a highly efficient 
  Rao-Blackwellized particle filter to generate grid maps from laser range data.
\end{itemize}

These experiments run on a turtlebot in real world which is connected with 
a four-core 8GB DRAM notebook as its computing unit. It runs a Linux
host OS with kernel 4.4.0 and ROS indigo. Since the turtlebot robot can not be shared
in current multi-robot frameworks, we ran the three workload one by one. 
For each experiment, we measured CPU utilization every 1s, for a period of 5 minutes at stable phases.

\begin{figure}[tb]
\centering
\includegraphics[width=8cm]{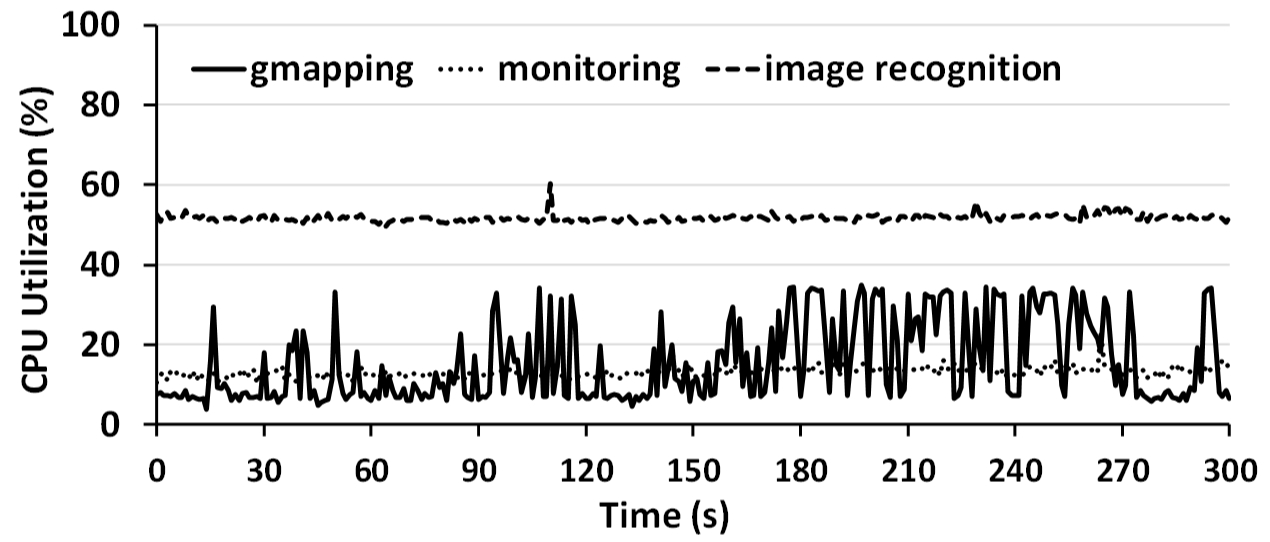}
\caption{Average CPU utilization over time of three robotics workloads. }
\label{fig:lowUtil}
\end{figure}

Figure \ref{fig:lowUtil} shows resource utilization for three robotics workloads. 
It's clear to see that current coarse-grained resource allocation approach
manages robots in an inefficient way:
The gmapping workload uses 15.3\% of CPU in the mapping process, while the monitoring workload occupies only 13.1\% of CPU
for image transfer. The image recognition is relatively better due to the 34-layer CNN consuming 51.7\% of CPU cycles.

According to these experiments, we can conclude that robot resources (i.e., CPU) are seriously underutilized with
current approaches. In next section, we will propose a new robotcenter OS Avalon to improve resource utilization 
by integrating two-level scheduling mechanism.

\section{AVALON OVERVIEW}
In this section, we first introduce the design overview of Avalon and present
the two challenges as well as our solutions. Finally, we will show
an illustrative example. 

\subsection{Design Overview}

\begin{figure}[tb]
\centering
\includegraphics[width=8cm]{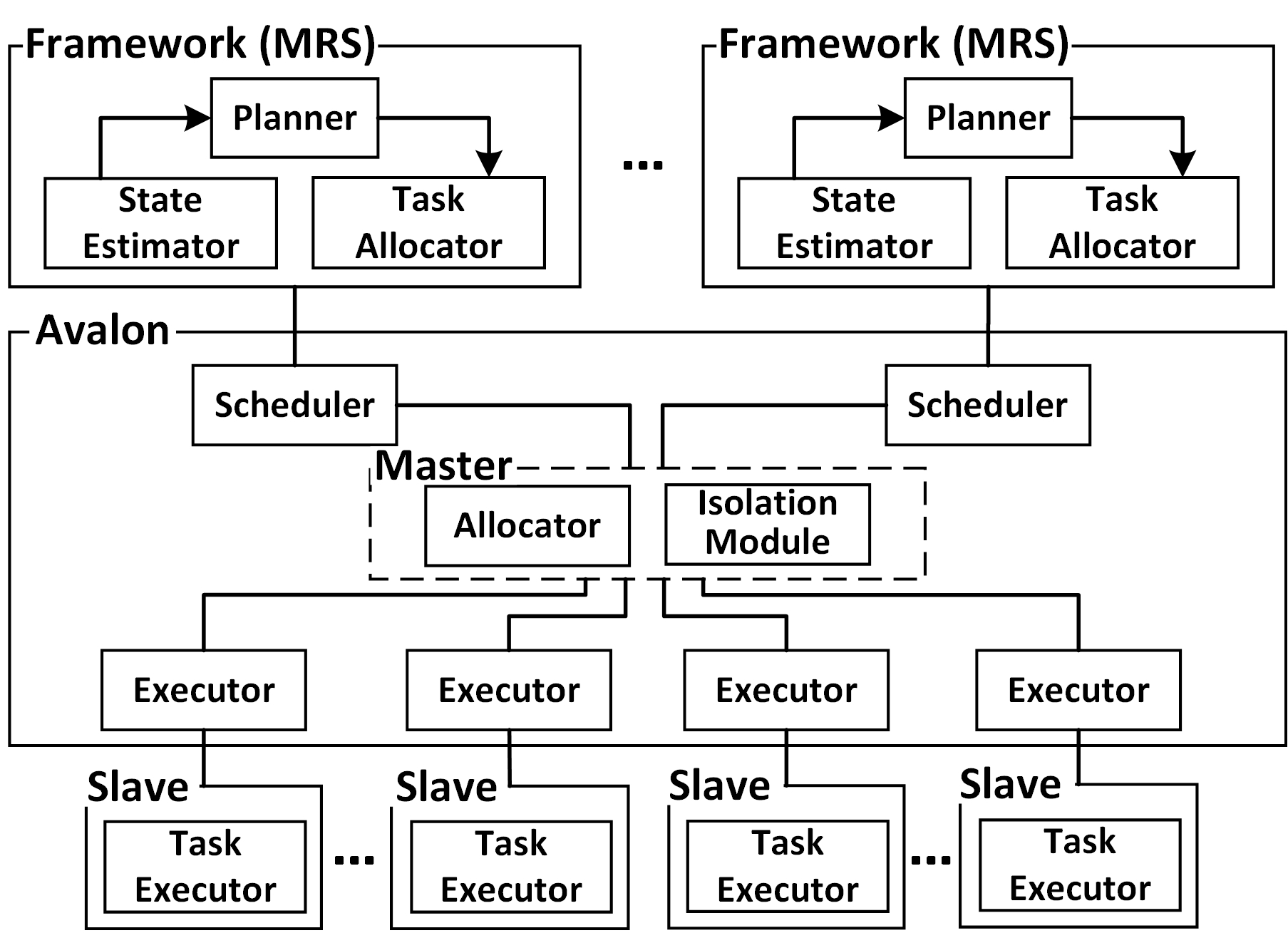}
\caption{Avalon architecture digram.}
\label{fig:arch}
\end{figure}

Figure \ref{fig:arch} shows the main components of Avalon. Avalon \emph{master} is a centralized task
controller that monitors and maintains the connection between multi-robot frameworks and robots.
Avalon supports multiple multi-robot frameworks, each of which requires to register a 
\emph{scheduler} into the master as a \emph{framework}. The Avalon master
determines how many available robots (called resource offers) to be pushed to a requesting framework based on position information. 
Once the framework accepts the offers, it launches an \emph{executor} process on robots to perform tasks. Executing process is
encapsulated into Linux Container \cite{LXC}, which is managed by \emph{Isolation Module}. Besides, each robot
runs a \emph{slave} daemon to communicate with the master and forward allocatable resources to the master periodically.

In such design, a robotcenter OS is divided into two levels: the master is at the lower level and responsible for
only fine-grained resource allocation; the frameworks at the higher level register a scheduler into
the master for task scheduling. According to this two-level scheduling design, 
robotics researchers could focus on developing specific multi-robot
frameworks without considering co-running multiple tasks. Furthermore,
it keeps Avalon simple and scalable to adapt to multi-tenant environment in robotcenters.

\subsection{Fine-grained Resource Sharing}

One challenge of the Avalon design is that a robot's resources have different 
features so that not all of them can be shared. To address this issue, 
we classify robot's resources into the following three categories:

\noindent
\textbf{Computation resources (CR)}: Some robots such as humanoid robots and self-driving vehicles 
are equipped with powerful cpu and high-capacity memory. 
This kind of resources can be shared by multiple tasks.

\noindent
\textbf{Sensory resources (SR)}: Some sensors including camera, laser and gps, 
  are the perceptual organs of robots to perceive the physical world. 
  This kind of resources are also shareable.

\noindent
\textbf{Action resources (AR)}: Actuators including wheel, hand and propeller, 
are the action organs of robots. Due to spacial limitations, this kind of 
resources are unshareable. For example, a robot cannot simultaneously move to 
position A and position B that are on opposite directions.

Based on this resource classification, we can categorize robotics workloads along three axes.
For instance, for the workloads mentioned in Section II, monitoring is an SR workload, 
image recognition is a CR-SR workload and gmmaping is an AR-CR-SR workload.
This gives a chance to deploy these three workloads on one robot in a mixed mode.

For each type of resource, Avalon uses two states \emph{used} and \emph{allocatable}  
to determine whether a resource can be shared. 
Usually a framework may run multiple tasks and one task can apply for multiple types of resource.
When resources are offered to a framework, the framework traverses 
the offer list to check whether the allocatable resources meet its requirement.
Specifically, Avalon provides interface for the framework to take the following actions:

\begin{itemize}
  \item If a task needs a CR execution environment, the allocated CR should be marked as \emph{used}, while
the remain CR of that robot will be marked as \emph{allocatable}.
  \item If a task needs an SR execution environment, all SR of that robot will be marked as \emph{allocatable}.
  \item If a task needs an AR execution environment, all AR of that robot will be marked as \emph{used}.
Because other AR may incur spatial conflicts with the used AR.
\end{itemize}

\subsection{Location-based Resource Offers}

Prior multi-robot systems are usually deployed for a specific
task with pre-allocated robots. In robotcenters, however, 
robots are allocated on-demand and dynamically, which raises
a significant challenge for robot resource allocation. 

Dynamic resource allocation in datacenters is also very challenging.
Datacenter OSes usually allocate resources and do scheduling based on 
data locality which means assigning computation tasks close to their data. 
However, the locality principle can not directly be employed in robotcenters 
because robots can move. Thus, Avalon needs a new resource allocation policy
to efficiently allocate resources to multiple frameworks.

To address this issue, we propose a location-based resource allocation policy. 
In particular, the Avalon master has a pluggable allocator 
module to offer a list of allocatable resources to frameworks.
The allocator module takes spatial constraint as an important utility. 
If available robots are far from the position where a framework task takes place, 
then the Quality of service (QoS) cannot be guaranteed 
since it takes long time for the robots to move from current positions to
the operation position.

To improve the efficiency of location-based allocation policy,
we implement a range \emph{filter} in Avalon allocator module (Algorithm \ref{alg:lbr}):
when a framework is registered with master, the selected operation position 
and search radius will be sent to the allocation module.
Avalon calculates the distance from each robot position to the 
framework operation position to form a score attached to related slave ID of a robot.
If robots locate in search area, it means that these robots are all allocatable for the framework.
The allocator module reorders the allocatable list by scores in an ascending order 
and then pushes several closest robots to the frameworks.

\begin{algorithm}[h]
\caption{Location based Reorder}
\label{alg:lbr}
\begin{algorithmic}
\REQUIRE
\STATE $R^t$    \COMMENT {A set of robots to be scheduled at time t}
\STATE $F^t$    \COMMENT {A set of frameworks at time t}
\STATE $pos^i$  \COMMENT {The position of ith robot}
\STATE $pos^j$  \COMMENT {The position selected by jth framework}
\STATE $r^j$      \COMMENT {the search radius selected by jth framework}
\ENSURE
\WHILE{$(F^t \neq 0)$}
\WHILE{$(R^t \neq 0$ and $F^j$ is available)}
\STATE calculate $score^i$ = distance($pos^i$, $pos^j$)
\IF{$(score^i \leq r^j)$}
\STATE push in list $l$
\ENDIF
\ENDWHILE
\STATE ascending reorder list $l$ based on score
\STATE offer list $l$ to $F^j$
\ENDWHILE
\end{algorithmic}
\end{algorithm}

We note that distance may not be a perfect utility in some complex environments 
because obstacles and speeds also affect allocation.
However, using distance is easy to implement and requires little computation. 
In addition, based on the two-level scheduling design philosophy, 
frameworks will choose suitable robots from the offer list by their own task schedulers.

\begin{figure}[tb]
\centering
\includegraphics[width=8cm]{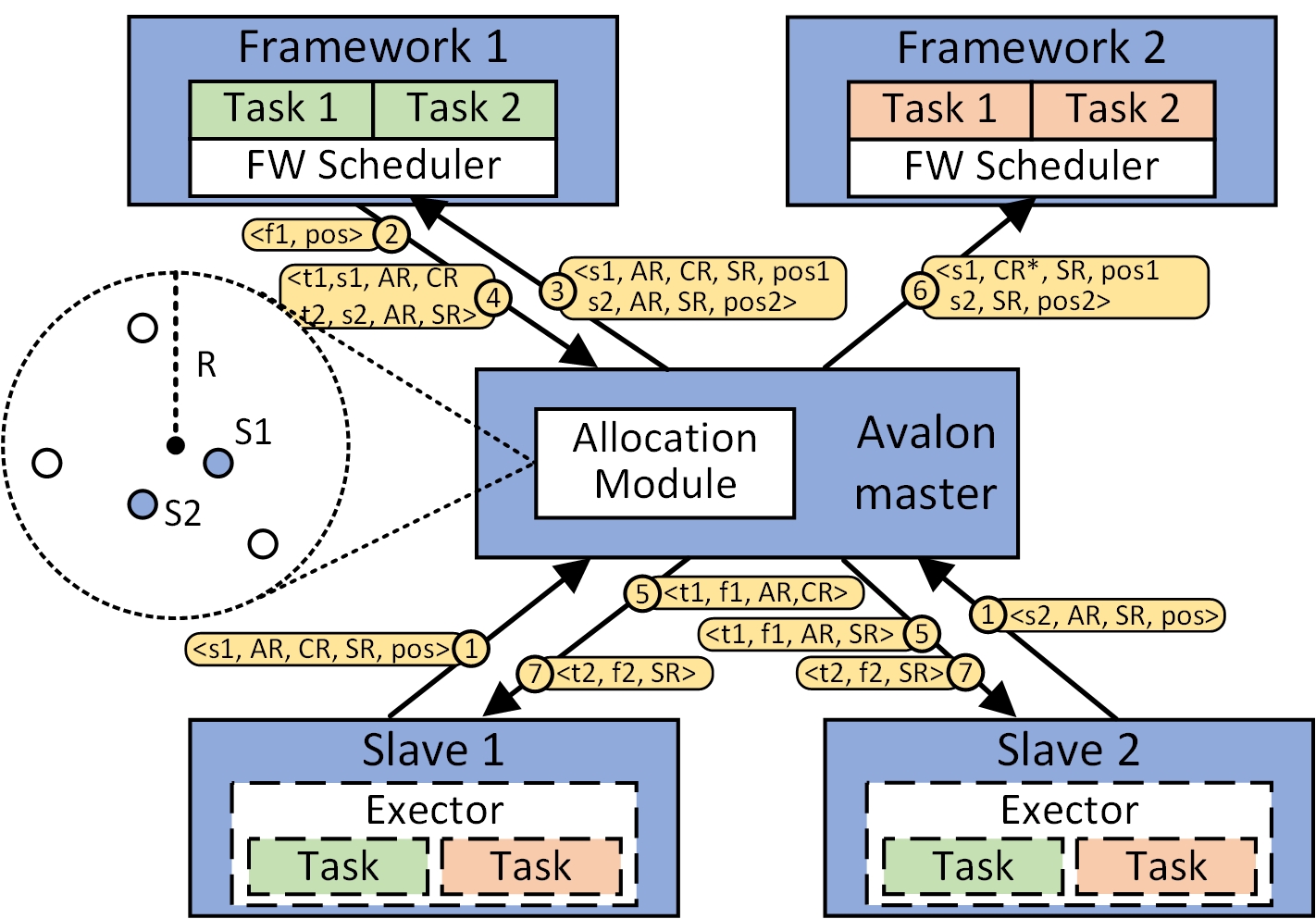}
\caption{Robots fine-grained sharing example. Assume two frameworks' select positions are same.}
\label{fig:offer}
\end{figure}

\subsection{Put It All Together}
Figure \ref{fig:offer} depicts an illustrative example of 
how Avalon fulfills the design of fine-grained resource sharing and 
location-based resource offers to allow resources shared by multiple frameworks.
The steps are shown as follow:

\begin{enumerate}
  \item Two robots (Slave 1 and Slave 2) report their available resources and position to the Avalon master;
  \item Framework 1 is registered with the master and sends the selected position and search radius to the master;
  \item The master sends a robot offer list to the framework based on algorithm \ref{alg:lbr};
  \item The scheduler of Framework 1 chooses proper robots from the offer list by its own task allocation policy;
  \item The master simultaneously deploys the task 1 and task 2 from Framework 1 to robot slave 1 and slave 2;
  \item Assume that the position and search radius selected by Framework 2 are the same with that of Framework 1, the master will offer the remain CR and all SR in slave 1 and all SR in slave 2 to Framework 2;
  \item Finally, the Avalon master deploys two tasks requiring SR in the two slaves.
\end{enumerate}

\section{IMPLEMENTATION}

We build an Avalon prototype on Linux with off-the-shelf ROS packages.
We use Linux containers to isolate each execution process.
Linux containers can reduce the complexity of configuring CPU quotas, memory limits and I/O rate limits,
which makes Avalon to easily scale up and down. 

We have implemented Avalon in C++ with an efficient actor-based asynchronous programming library called libprocess \cite{LibProcess}.
To allow Avalon to flexibly access multi-robot frameworks, our implementation 
exposes C++ and Python interfaces to bind applications with scheduler and executor (see Figure 3).
These two languages are well adapted in most ROS relevant applications.

Avalon has been evaluated in both simulation environment and real world. We will
open source Avalon soon. Next, we will present three implementation issues.

\subsection{Core Process}
All tasks are executed as processes in Linux containers. 
There are three types of core processes in Avalon (Figure \ref{fig:arch}): 
Master process, Slave process and Framework process. 
We will show how these three processes work as follows:

1) \emph{Master:} The master is the main controller module that polls events pool to trigger 
both frameworks and slaves. It maintains a dataset to record the status and resources of each robot (Slave). 
Once detecting a new framework is registered, the master will launch the allocator module 
to determine which slaves are available. In addition, the master also detects the executor 
status to allocate a next task or perform fault tolerance.

2) \emph{Slave:} The slave is installed in robots. Its responsibility is to monitor
 host robot's status and resources and communicate with the master.
 For computation resources (CRs), the slave detects allocatable CRs by system calls.
 For SR and AR, the slave obtains the information through robots themselves or user-input parameters. 
 The resource parameters have the following structure

\noindent
\texttt{--sr\underline{ }res="Sensor:{Function};..."}

\noindent
\texttt{--ar\underline{ }res="Actuator:{Function};..."}

\noindent
which is an unordered collection of key/value pairs. Note that the same key with different values would be
grouped into one tuple (eg. kinect:{ImageGen,LaserGen}). Furthermore, the slave obtains a robot's position 
from odometry topic published by ros node. Due to various topic names in different ros package, 
user should tells the slave module where to listen. 
In simulation environment, the slave also needs to know the namespace and coordinate transform between
map and each robot.

3) \emph{Framework:} The framework is responsible for choosing proper robots from a resource offer list and 
scheduling tasks on these robots. One framework usually has multiple tasks. To find proper robots, 
the framework should provide operation position and search range to the master. 
In a simulation environment, framework can select positions in the rviz GUI by 
publishing message into click point topic for visualization.

\begin{table}[tb]
\centering
\caption{Avalon API functions for schedulers and executors.}
\includegraphics[width=8.5cm]{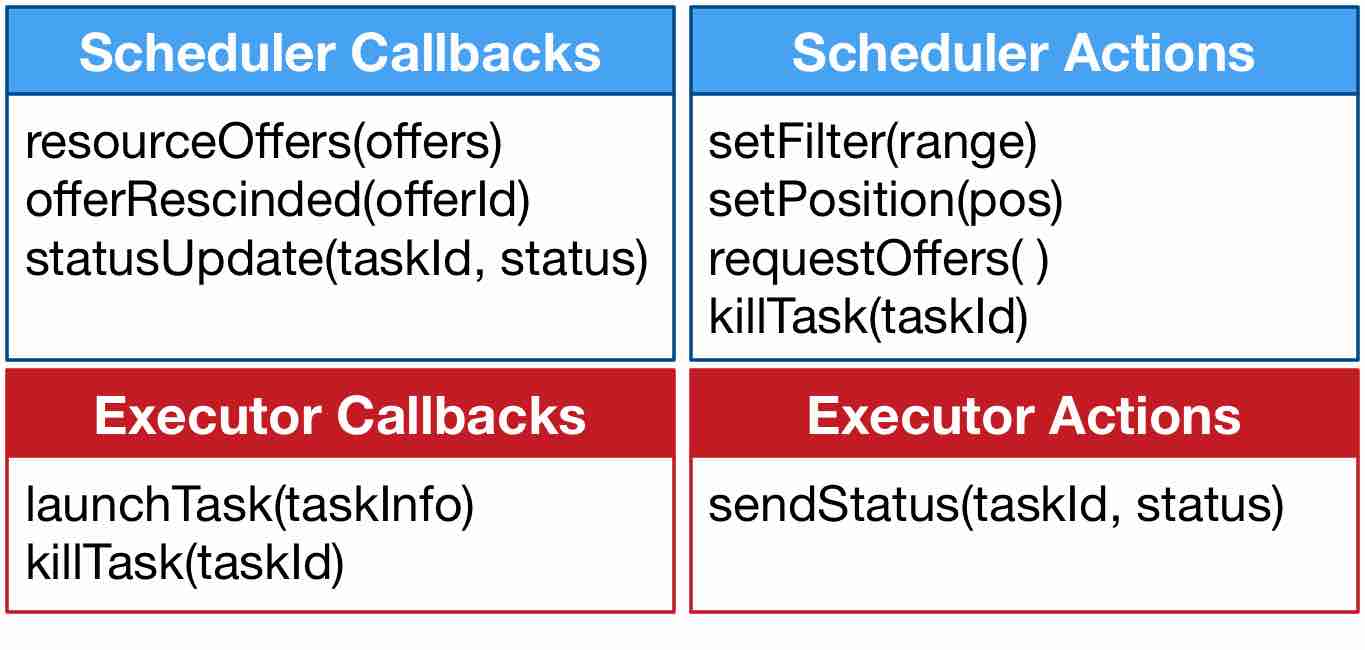}
\label{table:api}
\end{table}

\subsection{Access Interface}
Table \ref{table:api} summarizes the Avalon scheduler and executor API functions. The ``callback" columns list functions
that frameworks must implement and ``actions" columns list which they could invoke.

The \emph{Scheduler} interface binds functions with the state estimator and task allocator modules in multi-robot
framework in the Figure \ref{fig:arch}. Because a multi-robot framework may receive goals from multiple customers,
\emph{setFilter} and \emph{setPosition} could be invoked to update the framework parameters for different positions.
Then we implement \emph{resourceOffers} to allocate each task in one goal to specific robot after requesting offers.
The \emph{statusUpdate} monitors task status from slaves at runtime. If a framework does not respond an offer for 
a sufficiently long time, Avalon will \emph{rescind} the offer and reallocate the resources to other frameworks.

The \emph{Executor} interface binds functions with the task executor modules. Frameworks could directly control robots
through rostopic and rosservice API, or execute a roslaunch file in a container environment. The \emph{killTask} function
is used for a scheduler to kill one of its tasks. This will be useful when a robot encounters some emergency situations and
needs to stop its actions.

\subsection{Communication Protocols}
Avalon's communication protocols consist of two parts: \emph{internal communication protocol} and
\emph{external communication protocol}.
The internal communication protocol covers communication between Avalon processes.
It is implemented by Google's protocol buffers \cite{Protobuf} for flexibility and efficiency.
The external communication protocol defines the data transfer between slave-slave processes running in a ROS environment
and slave-framework processes.

Avalon uses \emph{scheduler} and \emph{executor} to communicate with a non-Avalon process running either on the slave or
in the framework. These two internal interfaces provide an asynchronous I/O mechanism to report initialized information of
framework and slave processes to master. Since tasks are executed in ROS context, executors provide converts to transform
a data message from the external communication format (serialized ROS message) to the internal communication format
(Protobuf message) and vice versa.

Meanwhile, Avalon also implements computing offloading through external communication protocol. Specifically,
cloud infrastructure can be registered with master as CR-only slaves, enabling multi-robot frameworks receive 
powerful computing resource pools from master. Since the computing environment is a ROS node, frameworks are able to launch multiple nodes within both servers and robots and coordinate with each other using ROS interprocess communication. It is worth noting that this approach can set up Cloud-Edge computing mode, which helps robots to offload heavy computation to local edge servers or remote cloud datacenters.

\section{EVALUATION}

We evaluated Avalon through two experiments on a turtlebot robot in real world.
Our first experiment aims at showing how one robot's resources
can be shared by three workloads to improve utilization.  Then we measured the communication overhead between two Avalon instances
running on different machines. The experimental results can help us to make decision for offloading in the Cloud-Edge computing mode.

\subsection{Resource Sharing}
\noindent
\textbf{Experimental settings: }
To evaluate the primary goal of Avalon, which is allowing diverse multi-robot frameworks to share a robot efficiently, we ran a mix of
three workloads described in Secection II: gmapping (CR-SR-AR), monitoring (SR) and image recognition (CR-SR). Specifically,
we launched three frameworks and the master in a local server (assume each workload is from different multi-robot frameworks).
A four-core 8GB RAM turtlebot is registered with the master as a CR-SR-AR slave, offering resources to each framework.
For performance metric, we measured CPU utilization for a period of 5 minutes at stable phases through perf tool \cite{Perf}.

\begin{figure}[tb]
\centering
\includegraphics[width=8cm]{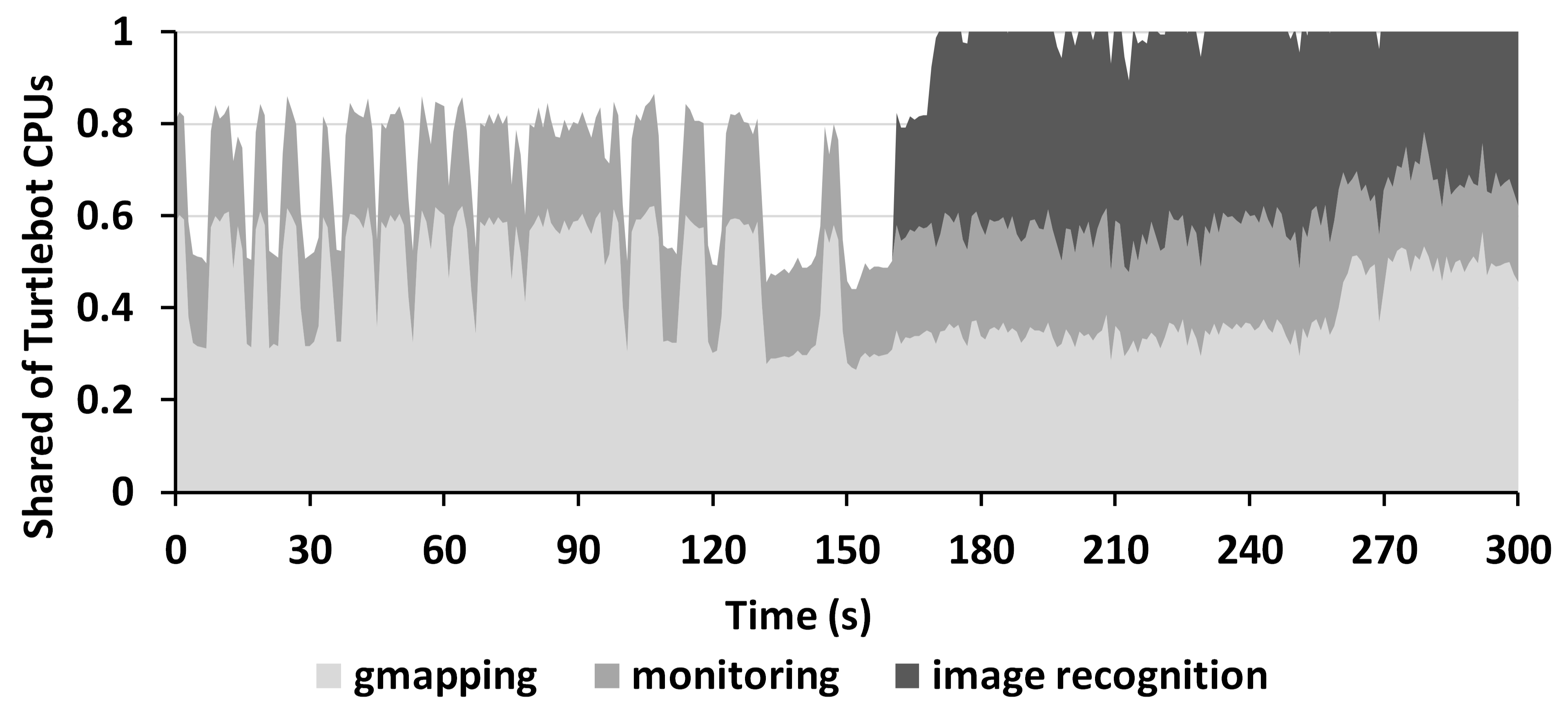}
\caption{Resource utilization with fine-grained resource sharing. }
\label{fig:sharing}
\end{figure}

\noindent
\textbf{Results: }
The effectiveness of Avalon would be proved by two facts:
Avalon achieves higher utilization than traditional multi-robot systems (MRSs) and 
workloads finish within acceptable time limits.
Our results show both effects, as detailed below.

Figure \ref{fig:sharing} shows the fraction of CPU running time allocated to each workload at different phases.
Assume both gmapping and image recognition workloads request [2CPU, 4GB RAM] computing resource.
At the first 165 seconds, turtlebot moves around and maps the physical world from laser range data.
Meanwhile, a monitoring workload receives image data from turtlebot's kinect sensor continuously, which improves
the CPU utilization up to 69.66\%.
After 165s, the third framework is registers with master and launches the image recognition workload into the turtlebot.
The CPU utilization of a mix of three workload increases to 98.38\% quickly.
Compared with the results with workloads in traditional single-task mode,
Avalon can improve the CPU utilization by 3.68 times (up to 7.51 times). 
We also show that the task finish time is acceptable in our video.

\subsection{Computation Offloading}
\noindent
\textbf{Experimental settings: }
We used the same configuration of the turtlebot from Sec. V-A and a local server with 4 cores and 8 GB RAM.
The connection between the turlebot and the local server was covered with a 2.4 GHz band wireless network in Beijing ICT CAS,
and remote servers were deployed in Tencent datacenter located in Guangzhou, China.
We show four communication paths analyzed for 6 data sizes as follows:
\begin{itemize}
  \item C2C: Container to container where both running in the same machine;
  \item R2LC: Container to container where one is hosted in the robot and the other in the local server;
  \item R2RC: Container to container where the other is executed in the remote cloud server;
  \item F2S: Communication between framework and slave which both hosted in the local server.
\end{itemize}
Note that C2C, R2LC and R2RC use an external communication protocol, which transfers string messages within ROS environment,
while F2S is implemented by an internal communication protocol.

\begin{figure}[tb]
\centering
\includegraphics[width=8cm]{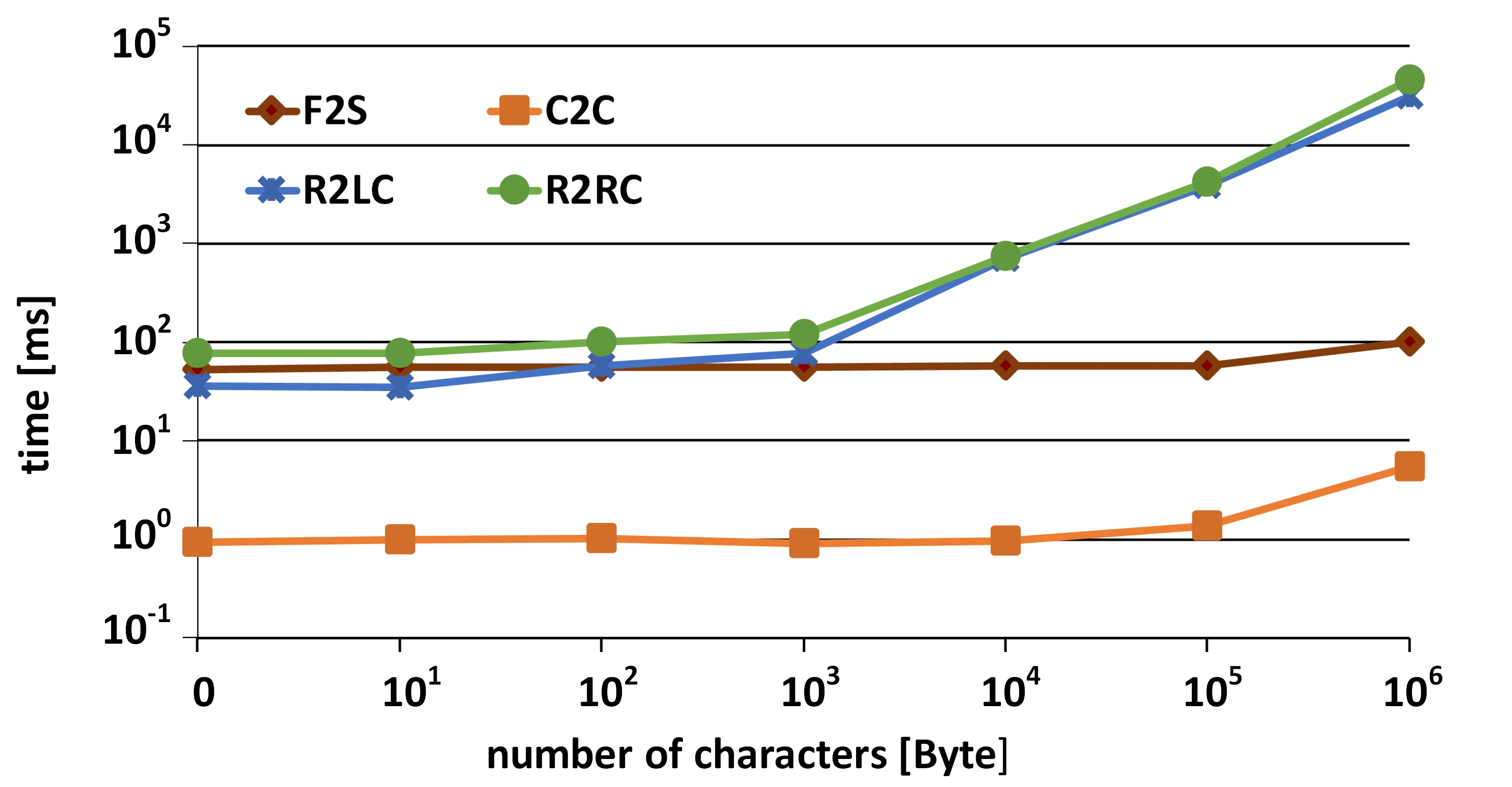}
\caption{RTT of four different type of communication paths in the Avalon system. }
\label{fig:latency}
\end{figure}

\noindent
\textbf{Results: }
Figure \ref{fig:latency} shows the round-trip time of four communication paths with 6 data sizes.
First, C2C is the smallest constraint of Avalon's throughput which takes lowest overhead shared by different frameworks.
Second, Avalon introduces an overhead of 100 [ms] for data sizes up to 1MB (see F2S in the figure), due to the
task queuing and internal communication overhead.
Third, external communication (R2LC and R2RC) is the biggest bottleneck of Avalon's throughput, which is caused by the variable
wireless network performance. When data size increases to 1MB, the overhead of R2RC brings a sharp rise. Thus, when  
determining which task can be offloaded to remote cloud, we should consider the package size and real time constrains of tasks.

\section{CONCLUSION AND FUTURE WORK}
This paper envisioned future robotcenter ecosystem that comprises multiple heterogeneous robots and proposed Avalon,
which is inspired by datacenter OS such as Mesos and Borg
Avalon is a two level scheduling robotcenter OS that enables multi-robot frameworks to share fine-grained resources.
We implemented Avalon on both Gazebo simulator and real world. Our experiments demonstrated that Avalon is able to achieve
high utilization through fine-grained resource sharing in robotcenter environment without substantial modifications 
in multi-robot frameworks. Moreover, Avalon supports offload computation intensive tasks into cloud servers, which 
makes sense to the Cloud-Edge computing mode.

Finally, we believe a robotcenter OS is needed for enabling the development of a future multi-robot ecosystem.
However, there are still a lot of open and challenging problems. We would like to researchers to
pay more attention to this field from operating system perspective.

\addtolength{\textheight}{-12cm}  










\begin{thebibliography}{99}

\bibitem{Kiva} KIVA, "KIVA System." [Online]. Available: www.kivasystems.com.
\bibitem{Swarmfarm} Swarmfarm, "Swarmfarm Robotics." [Online]. Available: http://www.swarmfarm.com
\bibitem{Heterogeneity} Z. Yan, N. Jouandeau and A. A. Cherif, "A Survey and Analysis of Multi-Robot Coordination," I.J.Robotics Res, vol. 10, no. 399, 2013.
\bibitem{RoboEarth} R. Janssen, R. Molengraft, H. Bruyninckx, M. Steinbuch, "Cloud based centralized task control for human domain multi-robot operations," Intelligent Service Robotics, vol. 9, no. 1, pp. 63-77 2016.
\bibitem{Rapyuta} D. Hunziker, M. Gajamohan, M. Waibel and R. D'Andrea, "Rapyuta: The RoboEarth Cloud Engine," ICRA, 2013, pp. 438-444.
\bibitem{Borg} A. Verma, L. Pedrosa, M. Korupolu, D. Oppenheimer, E. Tune and J. Wilkes, "Large-scale cluster management at Google with Borg," Eurosys, 2015, pp. 18:1-18:17.
\bibitem{Kubernetes} "Kubernetes Project," https://kubernetes.io/, 2017.
\bibitem{Mesos} B. Hindman, A. Konwinski, M. Zaharia, A. Ghodsi, A. D. Joseph, R. H. Katz, S. Shenker and I. Stoica, "Mesos: A Platform for Fine-Grained Resource Sharing in the Data Center," NSDI, 2011.
\bibitem{Hadoop} K. Shvachko, H. Kuang, S. Radia and R. Chansler, "The Hadoop Distributed File System," MSST, 2010, pp. 1-10.
\bibitem{Spark} M. Zaharia, M. Chowdhury, M. J. Franklin, S. Shenker and I. Stoica, "Spark: Cluster Computing with Working Sets," HotCloud, 2010.
\bibitem{Tensorflow} M. Abadi, P. Barham, et al, "TensorFlow: A System for Large-Scale Machine Learning," OSDI, 2016, pp. 265-283.
\bibitem{MapReduce} J. Dean and S. Ghemawat, "MapReduce: simplified data processing on large clusters," Commun. ACM, vol. 51, no. 1, pp. 107-113, 2008.
\bibitem{BSP} A. V. Gerbessiotis and L. G. Valiant, "Direct Bulk-Synchronous Parallel Algorithms," J. Parallel Distrib. Comput, vol. 22, no. 2 pp. 251-267 1994.
\bibitem{EdgeComputing} W. Shi, J. Cao, Q. Zhang, Y. Li and L. Xu, "Edge Computing: Vision and Challenges," IEEE Internet of Things Journal, vol. 3, no. 5 pp. 637-646 2016.
\bibitem{RS1} Y. Kim and D. A. Shell, "Distributed robotic sampling of non-homogeneous spatio-temporal fields via recursive geometric sub-division," ICRA, 2014, pp. 557-562.
\bibitem{RS2} J. Das, F. Py, J. B. J. Harvey, J. P. Ryan, A. Gellene, R. Graham, D. A. Caron, K. Rajan and G. S. Sukhatme, "Data-driven robotic sampling for marine ecosystem monitoring," I. J. Robotics Res, vol. 34, no. 12 pp. 1435-1452 2015.
\bibitem{Resnet} K. He, X. Zhang, S. Ren and J. Sun, "Deep Residual Learning for Image Recognition," CVPR, 2016, pp. 770-778.
\bibitem{LXC} "Linux Containers," 2012. [Online]. Available: http://lxc.sourceforge.net/
\bibitem{LibProcess} LibPorcess, "LibPorcess," [Online]. Available: http://www.eecs.berkeley.edu/~benh/libprocess
\bibitem{Protobuf} Google Inc, "protocol buffers," 2012. Available: [Online] http://developers.google.com/protocol-buffers
\bibitem{Perf} Perf Tool, "perf: Linux profiling with performance counters," 2015. Available: [Online] https://perf.wiki.kernel.org


\end{thebibliography}
\end{document}